\documentclass[sigconf]{acmart}

\AtBeginDocument{%
  \providecommand\BibTeX{{%
    \normalfont B\kern-0.5em{\scshape i\kern-0.25em b}\kern-0.8em\TeX}}}

\setcopyright{acmcopyright}
\copyrightyear{2022}
\acmYear{2022}
\acmDOI{XXXXXXX.XXXXXXX}

\usepackage{graphicx}
\usepackage{caption}
\usepackage{subcaption}
\usepackage{amsmath}
\usepackage{booktabs}
\usepackage{enumerate}
\usepackage{multirow}
\usepackage{color}

\usepackage[noend]{algpseudocode}
\usepackage{algorithmicx,algorithm}

\copyrightyear{2022}
\acmYear{2022}
\setcopyright{acmcopyright}
\acmConference[MM '22] {Proceedings of the 30th ACM International Conference on Multimedia }{October 10--14, 2022}{Lisbon, Portugal.}
\acmBooktitle{Proceedings of the 30th ACM International Conference on Multimedia (MM '22), October 10--14, 2022, Lisbon, Portugal}
\acmPrice{15.00}
\acmISBN{978-1-4503-9203-7/22/10}
\acmDOI{10.1145/3503161.3548217}

\begin{document}

\title{Generative Steganography Network}

\author{Ping Wei}
\email{pwei17@fudan.edu.cn}
\orcid{0000-0002-8852-6618}
\affiliation{%
	\institution{Fudan University}
	\streetaddress{Handan Road 220}
	\city{Shanghai}
	\country{China}}

\author{Sheng Li}
\email{lisheng@fudan.edu.cn}
\affiliation{%
	\institution{Fudan University}
	\streetaddress{Handan Road 220}
	\city{Shanghai}
	\country{China}}

\author{Xinpeng Zhang}
\authornote{Xinpeng Zhang and Zhenxing Qian are the corresponding authors.}
\email{zhangxinpeng@fudan.edu.cn}
\affiliation{%
	\institution{Fudan University}
	\streetaddress{Handan Road 220}
	\city{Shanghai}
	\country{China}}

\author{Ge Luo}
\email{18110240026@fudan.edu.cn}
\affiliation{%
	\institution{Fudan University}
	\streetaddress{Handan Road 220}
	\city{Shanghai}
	\country{China}}

\author{Zhenxing Qian}
\authornotemark[1]
\email{zxqian@fudan.edu.cn}
\affiliation{%
	\institution{Fudan University}
	\streetaddress{Handan Road 220}
	\city{Shanghai}
	\country{China}}

\author{Qing Zhou}
\email{21110240055@m.fudan.edu.cn}
\affiliation{%
	\institution{Fudan University}
	\streetaddress{Handan Road 220}
	\city{Shanghai}
	\country{China}}

\renewcommand{\shortauthors}{Ping Wei, et al.}

\begin{abstract}
   Steganography usually modifies cover media to embed secret data. A new steganographic approach called generative steganography (GS) has emerged recently, in which stego images (images containing secret data) are generated from secret data directly without cover media. However, existing GS schemes are often criticized for their poor performances. In this paper, we propose an advanced generative steganography network (GSN) that can generate realistic stego images without using cover images. We firstly introduce  the mutual information mechanism in GS, which helps to achieve high secret extraction accuracy. Our model contains four sub-networks, i.e., an image generator ($G$),  a discriminator ($D$), a steganalyzer ($S$), and a data extractor ($E$). $D$ and $S$ act as two adversarial discriminators to ensure the visual quality and  security of generated stego images. $E$ is to extract the hidden secret from generated stego images. The generator $G$ is flexibly constructed to synthesize either cover or stego images with different inputs. It facilitates covert communication by concealing the function of generating stego images in a normal  generator. A module named secret block is designed to hide secret data in the feature maps during image generation, with which high hiding capacity and image fidelity are achieved. In addition, a novel hierarchical gradient decay (HGD) skill is developed to resist steganalysis detection.  Experiments demonstrate the superiority of our work over existing methods.
\end{abstract}



\maketitle

\begin{figure}[htb]
	\centering
	\includegraphics[width=\linewidth]{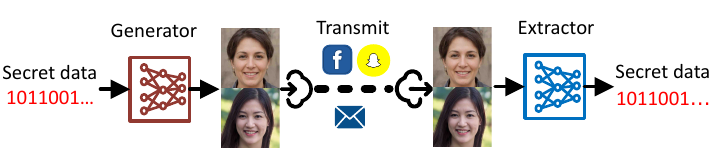}
	\caption{Schematic diagram of generative steganography. Input secret data is converted to natural stego images by a generator. Then these images are transmitted through lossless channels. The hidden secret data can be recovered from received stego images by an extractor.}
	\label{fig_flowchart}
\end{figure}

\section{Introduction}
\label{sec:intro}

Steganography is a technique that hides secret data in cover media for converting communication\cite{fridrich2009steganography}. Various types of cover media have been investigated for steganography\cite{survey_1}, including the digital audio \cite{yi19}, image \cite{tao19}, video \cite{xu2014data} and text \cite{borges2008robust}, where image is the most popular media for data hiding.  Early steganographic methods often modify the pixel values of the cover image's least significant bits to hide secret data\cite{hide}. Later, researchers paid more attention to the syndrome trellis coding (STC) framework\cite{filler2011minimizing, hill},  which aims to minimize the distortion caused by modifying cover images. These schemes can be called the carrier-modified-based methods, as cover images are modified to hide secret data.

Recently, a few deep learning (DL) based steganographic methods have been proposed\cite{survey_dl}. Tang \cite{tang2017automatic}  applies a network to learn the probability map of pixel alteration for data embedding. In \cite{hayes2017generating}, an adversarial training strategy is employed according to the 3-players game. Chu\cite{chu2017cyclegan} studies how CycleGAN\cite{cycleGAN} hides a source image into the generated images in an imperceptible way. Zhu\cite{zhu2018hidden} and Zhang\cite{ zhang2019steganogan} propose two novel networks to hide random binary secret data in the cover images. Baluja\cite{img_in_img} presents a system to hide color images inside another with minimal quality loss. Zhang\cite{udh} proposes a novel universal deep hiding architecture (UDH) to disentangle the encoding of secret images from the cover image. Yu\cite{yu2020attention} introduces the attention mechanism in the data hiding process. Jing\cite{hinet}  and Lu\cite{lu2021large} propose two reversible networks to hide secret images in a cover image. The schemes above all require cover images for data embedding. However, modifying the cover images will cause visual or statistical distortions, making the stego images easily detectable by steganalysis tools\cite{color_rich, ye2017deep,boroumand2018deep}. Once detected,  the behavior of covert communication fails.

To solve the problem, a new steganography manner called generative steganography (GS) emerges \cite{coverless}. Instead of modifying the cover image to embed secret data, it aims to synthesize stego images directly from secret data, as illustrated in Fig.\ref{fig_flowchart}. Cover images are not required in GS, thus, steganalysis tools will become ineffective. Several approaches\cite{otori2009texture, qian2017robust,Xu2015,li2018toward} have been proposed to synthesize some particular stego images to hide secret data, like  texture image\cite{otori2009texture, wkc2014} and fingerprint image\cite{li2018toward}. They can be called tailored GS methods for short. Nevertheless, only some special image types can be synthesized by them. Their data hiding capacities are much lower compared to traditional carrier-modified based works.

To make it more practical, researchers now use neural networks to generate natural stego images\cite{qin2020coverless}. They often map the secret data to the input labels\cite{liu2017coverless, zhang2020generative} or noise vectors\cite{hu2018novel, hu-2} of GANS by a pre-built mapping rule. Then stego images can be generated with the assigned noise vectors or labels according to the mapping rule. Data receivers can extract the hidden secret data from received stego images by a pre-trained extractor. For simplicity, we term such schemes as deep learning (DL) based GS solutions. Compared with tailored GS schemes, these DL-based GS methods can generate more natural stego images. However, their data hiding capacities are much lower due to the limitation of label number and noise vector size. Meanwhile, stego images generated by these methods are often visually poor. Transmitting such images may arouse suspicion of the monitor in covert communication.

In this paper, we propose a novel generative steganography network (GSN) integrated with mutual information mechanism. To disguise transmitting secret messages through stego images, we propose a flexible image generator that can flexibly synthesize cover images or stego images according to the inputs. The main contributions of our work are:
\begin{enumerate}[1)]
\item We propose a holistic steganography solution for covert communication, in which mutual information is first introduced in generative steganography. 

\item We hide the function of generating stego images in a flexibly constructed generator,  which can generate either cover or stego images depending on the inputs. 

\item A novel technique called hierarchical gradient decay (HGD) is proposed to improve the steganalysis resistance. 

\item Proposed GSN achieves better performances than state-of-the-art works. It can synthesize realistic stego images with high hiding capacity, secret extraction accuracy, and security.

\end{enumerate}

\section{Related Works}
Most of the published steganographic works are carrier-modified based solutions that require cover images\cite{survey_dl, img_in_img}. But this paper focuses on generative steganography (GS) that doesn't require cover images. Existing GS schemes can be roughly classified into two categories: tailored GS and deep learning (DL) based GS.

\textbf{Tailored generative steganography} In these methods, stego images are synthesized according to some handcrafted procedures, where the secret data is encoded into specific textures or patterns. Otori\cite{otori2009texture} first proposes encoding secret data into dotted patterns, then stego texture images are synthesized by painting these patterns. In \cite{wkc2014}, the authors propose a secret-oriented texture synthesis solution, where secret data is embedded by pasting proper source texture on different locations of the synthesized images, referring to an index table. In \cite{Xu2015}, the authors suggest using marbling images to hide secret data. Secret messages are printed on the background of an image, and then this image is deformed into different marbling patterns using reversible functions. In \cite{li2018toward}, secret data is encoded as the positions and polarities of minutia points in fingerprint images. Then, stego fingerprint images can be constructed using a phase demodulation model using these encoded minutiae positions. The main disadvantage of these tailored GS schemes is that their generated stego image contents are unnatural, which is unsafe or even suspicious in covert communication.  

\textbf{DL-based generative steganography} To produce natural stego images,  some researchers propose to synthesize images with networks. Works\cite{chu2017cyclegan, duan} use GANs to transform secret images into meaning-normal stego images, and these stego images can be converted back with another image generator. But they cannot convey random secret data as other steganographic methods do. It is more practical to transmit secret data in steganography. Therefore, Liu\cite{liu2017coverless} and Zhang\cite{ zhang2020generative} propose to map binary secret data to the class labels of ACGAN\cite{odena2017conditional}. By which, different stego images can be generated with the corresponding labels according to the given secret data and mapping rule. The labels of stego images can be extracted using a classifier, and then hidden secret data is recovered according to the mapping rule. Similarly, Hu\cite{hu2018novel} establishes a mapping rule between the secret data and the input noise vector of DCGAN\cite{radford2015unsupervised}. Then stego images can be generated with the mapped noise values per the mapping rule and given secret data. In work \cite{GSS}, the authors propose an image inpainting-based GS solution. Where secret messages are embedded in the remaining region of a corrupted image with Cardan grille, then this image is fed into a pre-trained generator for stego image generation. Wang\cite{sstegan} proposes to generate stego images from the concatenation of binary secret data and noise vector using GANS, in which secret data is input to the generator directly. Though the DL-based GS schemes can produce realistic stego images, they are rather rudimentary with poor performance. The stego images generated by them are often of low visual quality. In addition, their hiding capacities are often limited to several hundred bits per image. They can not be improved with the increase of stego image size due to the limitation of GANs' input dimensions. For example, only 6 bits of secret data are conveyed by every 28×28 stego image in work \cite{liu2017coverless}, and each 32×32 stego image carries 400 bits of secret data in \cite{sstegan}.
 
\section{Proposed Method GSN}
\begin{figure}[t]
	\centering
	\includegraphics[width=\linewidth]{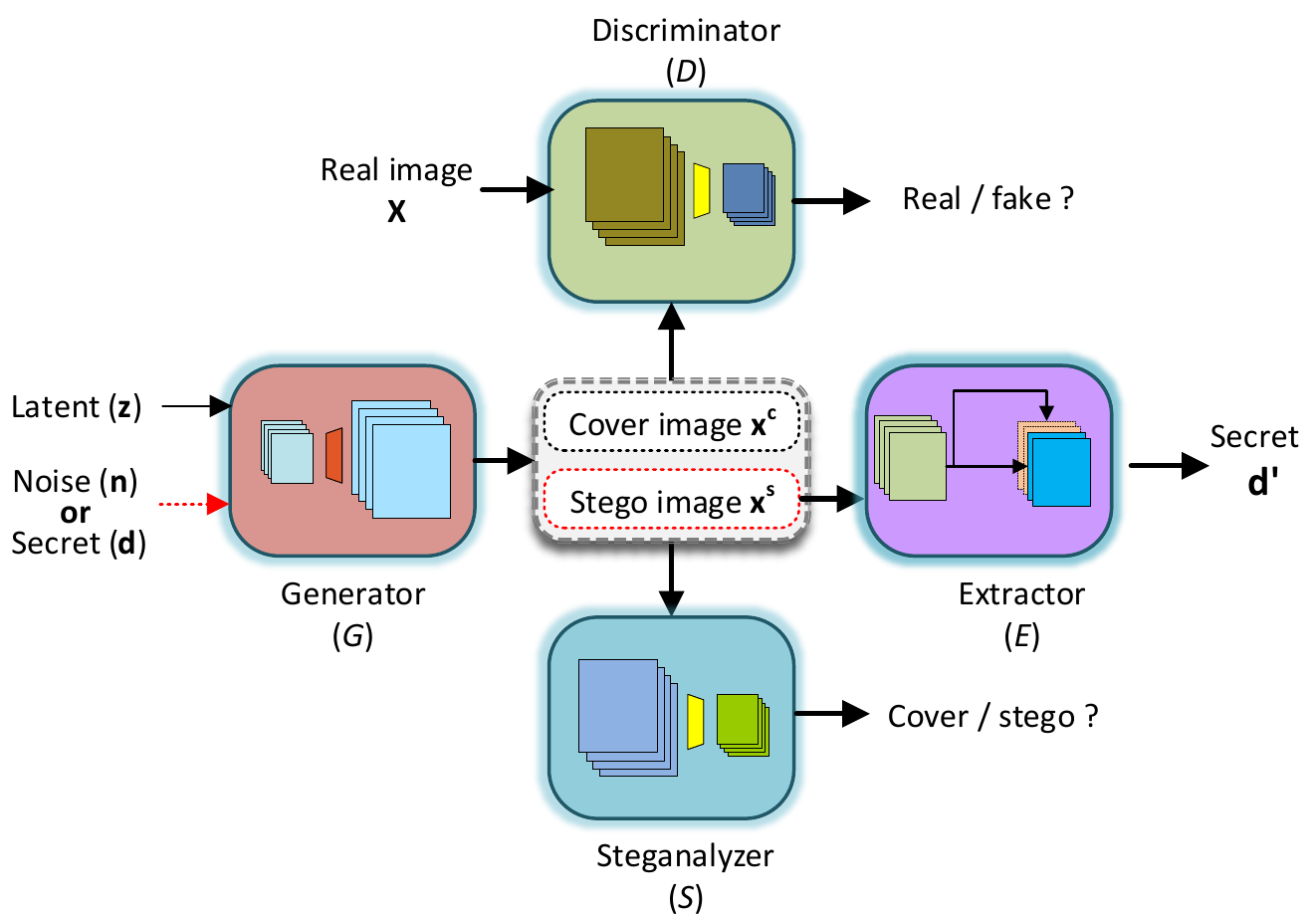}
	\caption{The overall framework of our proposed GSN. A cover/stego image can be generated when (z, n)/(z, d) is input. \textit{D} and \textit{S} act as dual discriminators to ensure the visual quality and statistical imperceptibility of cover/stego images. \textit{E} aims to reveal the hidden secret from generated stego image. }
	\label{fig1}
\end{figure}

The architecture of the proposed GSN is given in Fig.\ref{fig1}, which consists of a generator (\textit{G}), a discriminator (\textit{D}), a steganalyzer (\textit{S}) and an extractor (\textit{E}). \textit{D} and \textit{S} are used as two discriminators in GANs, which can ensure the visual quality and reduce the difference between generated cover/stego images, respectively. The inputs of GSN include a latent vector \textbf{z}, a noise matrix \textbf{n}  or a three dimensional matrix of secret data \textbf{d}. The generator can produce either a cover image $\mathbf{x^c}$ or stego image $\mathbf{x^s}$, depending on which of (\textbf{z, n}) and (\textbf{z, d}) is input. Then, real image and the generated stego image are sent to the discriminator to decide whether they are real or fake. Meanwhile, the generated cover/stego images are fed into the steganalyzer for differentiation. The generated stego images are input to the extractor, and  $\mathbf{d}'$ is the predicted secret.

\subsection{Problem Formulation}
In our scheme, a stego image can be generated with secret data \textbf{d} and latent \textbf{z}, i.e, $\mathbf{x^s}=G(\mathbf{z},\mathbf{d})$. Secret data  influences the image content, which should be recovered exactly from the generated stego image $\mathbf{x^s}$. From the perspective of information theory, the  mutual information between \textbf{d} and $\mathbf{x^s}$ is expected to be maximized, i.e., $ \mathrm{max} ~ I (\mathbf{d}, G(\mathbf{z},\mathbf{d})) $. That is to say, input secret data and the generated stego images are closely related. Different stego images should be generated when the input secret data is varied, and the hidden secret is hoped be extracted accurately from  generated stego images. Therefore, we incorporate the mutual information into GANs for data hiding. The loss function can be defined as:
\begin{align}
	\label{eq2}
	\min _{G} \max _{D} [\mathcal{L}_{adv}(D, G)-\lambda \cdot I (\mathbf{d}, G(\mathbf{z},\mathbf{d}))], 
\end{align}
where,  $\mathcal{L}_{adv}(D, G)$ is the adversarial loss of GANs, i.e, $\mathcal{L}_{adv} = 	\mathbb{E}_{\mathbf{x} \sim p_{x}} logD(\mathbf{x}) + \mathbb{E}_{\mathbf{z} \sim p_{z}, \mathbf{d} \sim p_{d}} log[  1-D(G(\mathbf{z},\mathbf{d}))] $. Here, $\mathbb{E}$  stands for the expectation. $I (\mathbf{d}, G(\mathbf{z},\mathbf{d}))$ is the mutual information between secret data and the generated stego image. \textit{G} wants to minimize while \textit{D} expects to maximize this loss function. 

But mutual information $I (\mathbf{d}, G(\mathbf{z},\mathbf{d}))$ is hard to be achieved as it requires the posterior distribution $ p(\mathbf{d} \mid  G(\mathbf{z},\mathbf{d})) $. Inspired by Infogan\cite{infogan},  a variation lower bound $ \mathcal{L}_d{(G, E)} $ is used to approximate $I (\mathbf{d}, G(\mathbf{z},\mathbf{d}))$:

\begin{equation}
\begin{aligned} 
\label{eq3}
&I(\mathbf{d}, G(\mathbf{z}, \mathbf{d})) \\ 
&=H(\mathbf{d})-H(\mathbf{d} \mid G(\mathbf{z}, \mathbf{d})) \\ 
&= H(\mathbf{d}) + \mathbb{E}_{\mathbf{x^s} \sim G(\mathbf{z}, \mathbf{d})}\mathbb{E}_{\mathbf{\hat{d}} \sim p(\mathbf{d} \mid \mathbf{x^s})}\log p(\mathbf{\hat{d}} \mid \mathbf{x^s}) \\ 
&= H(\mathbf{d})+\mathbb{E}_{\mathbf{x^s} \sim G(\mathbf{z}, \mathbf{d})}[\underbrace{D_{\mathrm{kl}}(p(\cdot | \mathbf{x^s}) \| q(\cdot | \mathbf{x^s}))}_{\geq 0} + \mathbb{E}_{\mathbf{\hat{d}} \sim p(\mathbf{d}\mid \mathbf{x^s})} \log q(\mathbf{\hat{d}} | \mathbf{x^s})] \\ 
& \geq H(\mathbf{d}) +  \mathbb{E}_{\mathbf{x^s} \sim G(\mathbf{z}, \mathbf{d})}\mathbb{E}_{\mathbf{\hat{d}} \sim p(\mathbf{d} \mid \mathbf{x^s})}\log q(\mathbf{\hat{d}} \mid \mathbf{x^s}) \\
&= H(\mathbf{d})  + \mathbb{E}_{\mathbf{d} \sim p_d, \mathbf{x^s} \sim G(\mathbf{z},\mathbf{d})}\log q(\mathbf{d} \mid \mathbf{x^s})  \\
&=\mathcal{L}_{d}(G, E). \\
\end{aligned}
\end{equation}

In fact, $\mathcal{L}_{d}(G, E)$  can be computed by \textit{G} and \textit{E}:
\begin{equation}
\max _{G, E}\mathcal{L}_{d}(G, E) = H(\mathbf{d})  + \mathbb{E}_{\mathbf{d} \sim p_d, \mathbf{z} \sim p_z} \mathbf{d} \cdot \log [E(G(\mathbf{z, d}))], 
\end{equation}
where,  $ H(\mathbf{d}) $ is the entropy of secret data, which has a constant value. $ D_{\mathrm{kl}}(\cdot)$ is the KL divergence. $\mathbf{d} \sim p_d$ means sampling secret data \textbf{d} from  distribution $p_d$ ($ \sim B(n ,0.5)$). $\mathbf{z} \sim p_z$ means sampling latent \textbf{z} from normal distribution $p_z$ ($ \sim N(0,1)$). $\mathbf{x^s} \sim G(\mathbf{z}, \mathbf{d}) $ means sampling stego images. $ q(\mathbf{d} \mid \mathbf{x^s}) $ is an auxiliary distribution used to approximate the true posterior probability $ p(\mathbf{d} \mid  G(\mathbf{z},\mathbf{d})) $. $\mathcal{L}_{d}(G, E)$ can be calculated with the Monte Carlo simulation: randomly sample a secret tensor \textbf{d} and  a noise tensor \textbf{z} to synthesize a stego image $\mathbf{x^s}$ using the generator \textit{G}, and then extract the hidden secret with extractor \textit{E}. Both \textit{G} and \textit{E} hope to maximize the lower bound. $E(G(\cdot))$ means extracting the hidden secret from generated stego image, the values of which fall in (0, 1) after a Sigmoid operation.  Only when the extracted secret is equal to the  input binary secret data \textbf{d}, the maximum value of $ \mathcal{L}_d{(G, E)} $ is reached, and thus mutual information $I (\mathbf{d}, G(\mathbf{z},\mathbf{d}))$ is maximized. 

In our scheme, \textit{G} can synthesize cover image ($\mathbf{x^c}=G(\mathbf{z},\mathbf{n})$) and  stego image  ($\mathbf{x^s}=G(\mathbf{z}, \mathbf{d})$). A steganalysis algorithm SR-net\cite{boroumand2018deep} is used as the backbone of steganalyzer \textit{S}, which aims to minimize the statistical difference between generated cover/stego images. The adversarial loss between \textit{G} and \textit{S} is written as  $\mathcal{L}_{S}(S, G)$. Different to  $\mathcal{L}_{d}(G, E)$ , \textit{S} hopes to output the correct predictions ([0,1]or[1,0]) with an binary cross-entropy loss, while \textit{G} wants \textit{S} to output [0.5, 0.5] for both cover and stego images, as described in Eq.\ref{eq-lsteg} and Eq.\ref{eq-entropy-loss}. Both \textit{G} and \textit{S} aim to minimize these two losses. 

To generate realistic stego images with high secret extraction rate and good undetectability, we combine the loss functions above and set the overall optimization object as:
\begin{equation}
	\label{total_loss}
	\begin{aligned}
		&\min_{(G, S, E)} \max_{D} \mathcal{L}_{total}(D, G, S, E) \\ &= \mathcal{L}_{adv}(D, G) - \lambda \cdot \mathcal{L}_{d}(G, E) + \beta \cdot \mathcal{L}_{S}(S,G), 
	\end{aligned}
\end{equation}
here, both \textit{G}, \textit{S} and \textit{E} expect to minimize $L_{total}(D, G, S, E)$,  while \textit{D} wants to maximize it. $\lambda$ and $\beta$ are two hyper-parameters. 

\subsection{Loss Functions}
\label{loss_functions}
In this section, we decompose $L_{total}(D, G, S, E)$ into specific loss functions for each sub-network.

\begin{figure}[htb]
	\centering
	\includegraphics[width=0.9\linewidth]{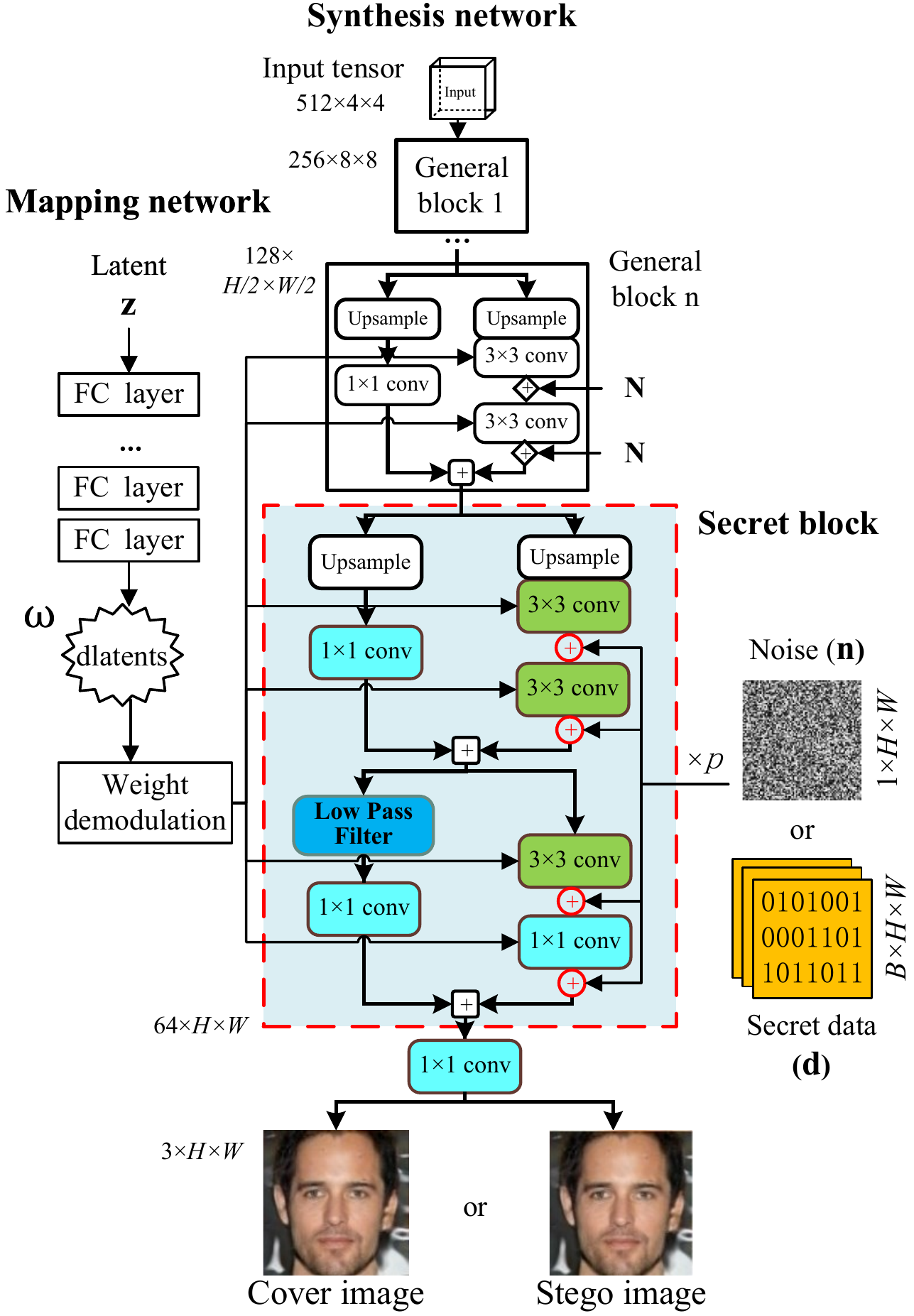}
	\caption{The architecture of proposed generator. The generator can synthesize either cover or stego image, when secret block is input with \textbf{n} or \textbf{d}, respectively.}
	\label{fig2}
\end{figure}

\noindent \textbf{Generator's loss}
The loss of generator takes two adversarial training processes and a regularization item into consideration:
\begin{equation}
	\label{eq-total-loss-g}
	Loss_{G} = \mathcal{L}_{adv} + \lambda_{1}\cdot \mathcal{L}_{steg} + \lambda_{2} \cdot \mathcal{R}_{PL},  \qquad
\end{equation}
\begin{align}
	\mathcal{L}_{adv} = & ~ \mathbb{E}_{\mathbf{z} \sim p_{z}, \mathbf{d} \sim p_{d}} log[  1-D(G(\mathbf{z},\mathbf{d}))]  \nonumber  \\ & +  
	\mathbb{E}_{\mathbf{x} \sim p_{x}} logD(\mathbf{x}), \\
	\mathcal{L}_{steg} = & ~ \mathbb{E}_{\mathbf{z} \sim p_{z}, \mathbf{n} \sim p_n, \mathbf{d} \sim p_{d}}
	\|0.5-S(G(\mathbf{z},\mathbf{n}))\|_2 \nonumber \\
	\label{eq-lsteg}
	& + \| 0.5-S(G(\mathbf{z},\mathbf{d}))\|_2, \\
	\mathcal{R}_{PL} = &  ~ \mathbb{E}_{{\omega,\psi} \sim N(0,1)} (\| J_{\omega}^{T} \|_{2} - \eta), 
\end{align}
here, $\mathcal{L}_{adv}$ is the adversarial loss between $D$ and $G$. $\mathcal{L}_{steg}$ is the adversarial loss of \textit{G} against \textit{S}, which ensures the outputs of $S$ are close to 0.5 for both cover and stego images (i.e., $S$ could not distinguish the source of images). $\mathcal{R}_{PL}$ is a regularization item used in \cite{karras2020analyzing} to  improve the training stability and disentangle the dlatents space.  $\lambda_{1}$ and  $\lambda_{2}$ are two hyper parameters;  $p_{z}$, $p_{n}$ and  $p_x$ refer to the distributions of latent, noise and real image; $\psi$ is a randomly generated noise image with normal distribution, and $\omega$ is a dlatents variable as shown in Fig.\ref{fig2}; $ \| J_{\omega}^{T} \|_{2} $ is the Jacobian matrix of $\psi$ with respect to $\omega$, and $\eta$ is a weighted value calculated with mean  $\| J_{\omega}^{T} \|_{2}$; $\|\cdot\|_2$ means L-2 norm.

\noindent \textbf{Discriminator's loss}
Discriminator's loss is defined as:
\begin{align}
	\label{eq-loss-D}
	&Loss_{D}=-\mathcal{L}_{adv} + \alpha \cdot \mathcal{R}_{1}, \\
	&\mathcal{R}_{1} = \frac{\beta}{2}\mathbb{E}_{\mathbf{x}\sim p_{x}}(\| \nabla D(\mathbf{x}) \|^{2} ) \nonumber,
\end{align}
here, $\mathcal{L}_{adv}$ is the adversarial loss as Eq.\ref{eq-total-loss-g}. $\alpha$ is a hyper parameter and $\mathcal{R}_1$ is the R1 regularization item given by \cite{mescheder2018training}; $ \| \nabla D(\mathbf{x}) \|^{2} $ means the squared gradient of discriminator's output $D(\mathbf{x})$ with respect to input real image \textbf{x}, and $\beta$ is a constant.

\noindent \textbf{Steganalyzer's loss}
We adapt the binary cross-entropy loss in steganalyzer (\textit{S}). \textit{S} outputs a two-dimension vector rather than a scalar like GANs, which is trained to output the correct predictions([0, 1] or [1, 0]) for input cover/stego images. 
\begin{equation}
	\begin{split}
		Loss_{S} =& -\mathbb{E}_{\mathbf{z}\sim p_z,\mathbf{n}\sim p_n, \mathbf{d}\sim p_d} [\mathbf{y}_{1} \cdot  logS(G(\mathbf{z},\mathbf{n})) \\	&	+  \mathbf{y}_{2} \cdot logS( G(\mathbf{z},\mathbf{d}))],
	\end{split}
	\label{eq-entropy-loss}
\end{equation}
here, $\mathbf{y}_{1}$/$\mathbf{y}_{2}$ refers to the ground truth of cover/stego image. 

\noindent \textbf{Secret extraction loss}
The loss of extracting hidden secret is computed by the binary cross entropy, with the prediction result $\mathbf{F}$ (as shown in Fig.\ref{fig_extr}), added noise and the input binary data \textbf{d}:
\begin{align}
	Loss_{E}&=-\mathbb{E}_{\mathbf{z}\sim p_{z}, \mathbf{d}\sim p_{d}} \mathbf{d} \cdot log [E(G(\mathbf{z, d})+ noise)] \nonumber \\
	&=-\mathbb{E}_{\mathbf{z}\sim p_{z}, \mathbf{d}\sim p_{d}} \mathbf{d} \cdot log[Sigmoid(\mathbf{F} + noise)],
	\label{eq-secret-loss}
\end{align}
where, $p_{d}$ denotes the Bernoulli distribution of binary secret data. $Sigmoid(\cdot)$ force the results to fall into (0, 1). Here, we add random noise ($\sim N(0, 0.01)$) to generated stego images for improving the robustness.  Both \textit{G} and \textit{E} are optimized to minimize this loss. 

\subsection{Training strategy}
To effectively train our GSN, sub-networks $G$, $D$, $S$ and $E$ are optimized sequentially as illustrated in Algorithm \ref{alg1}.  $G$ and $E$ are optimized simultaneously to improve the secret extraction accuracy, where a hierarchical gradient decay (HGD) skill (will be introduced in Sec.\ref{HGD})  is applied to improve the resistance against steganalysis methods. We optimize \textit{G} with $Loss_{G}$ and $Loss_{E}$ separately, mainly to decrease the differences between generated cover/stego images. The real images are only used to train \textit{D}. 
\begin{algorithm}[t]
	\caption{Training strategy}
	\label{alg1}
	\begin{algorithmic}[1]
		\Require  A set of real images, secret data \textbf{d}, noise \textbf{n}
		\Ensure  The trained GSN model
		\For{each \textit{step}}
		\State Generate \textit{m} pairs of cover and stego images with $G$;
		\State optimize $G$ to minimize $Loss_{G}$;
		\State optimize $D$ to minimize $Loss_{D}$;
		\State optimize $S$ to minimize $Loss_{S}$;
		\State optimize $G$ + $E$ to minimize  $Loss_{E}$,  apply HGD skill in $G$. 
		\EndFor
	\end{algorithmic}
\end{algorithm}

\subsection{Structure of Generator}
To facilitate covert communication, we propose a flexible image generator that can produce either cover or stego images. Fig.\ref{fig2} illustrates the architecture of proposed generator. It is improved based on stylegan2\cite{karras2020analyzing}, which consists of a mapping network and a synthesis network. The former network maps the input latent vector  \textbf{z} into an intermediate dlatents vector $\omega$ using eight fully connected (FC) layers. Next, $\omega$ is passed through a module called weight demodulation, which controls the image style. The synthesis network contains several general blocks and a newly designed secret block. The input of \textit{G}  is a 512×4×4 trainable tensor initialized with normal distribution. Higher resolution feature maps can be generated after up-sampling. The general blocks are made up of an upsample operation and several convolution layers. Noise matrix \textbf{N} ($\sim N(0, 1)$, shape: $1 \times h \times w$) is added to the feature maps.

We design a module called secret block to enable the generator to synthesize cover and stego images, as shown in Fig.\ref{fig2}. It contains three 1×1 convolution layers (followed by a LeakyRelu operation), three 3×3 convolution layers (with LeakyRelu), two upsample operations and four data merging operations (denoted as red \textcolor{red}{$\oplus$}) and a low pass filter. The weight demodulation module regulates each convolutional layer to adjust the image content. Here, we add a noise matrix \textbf{n} ( $\sim N(0, \sigma)$, shape: $1 \times \textit{H} \times \textit{W}$ ) into feature maps to generate cover images as original stylegan2, and add a three-dimension secret matrix \textbf{d} ($ \sim B(n, 0.5)$, $\textit{B} \times \textit{H} \times \textit{W}$ ) to generate stego images. The generator can synthesize common cover images without secret as stylegan2, or it can be used to generate stego images when necessary. After training, realistic cover/stego images are generated, which can hardly be distinguished by eyes or steganalysis tools. The sender can pretend to transmit synthetic cover images while sending secret messages through stego images, which improves covert communication security.

The function of the secret block lies on two aspects: 1) it enables the generator to synthesize cover or stego images when noise or secret data is input; 2) it removes the defects on image contents and improves the quality of stego images. The input secret data  can be written as  $\mathbf{d}=\left\{0,\ 1\right\}^{B \times H\times W}$. Here $H/W$ is the height/width of \textbf{d} , which is of the same size as output stego images. $B$ refers to the channel number, which determines the payload of generated stego images. In the secret block, convolutional feature maps of each layer can be expressed as $\mathbf{F}=\{\mathbf{f}_i\}|_{i=1}^N$,  here \textit{N} denotes the total channel number of feature maps. The data merging  operation \textcolor{red}{$\oplus$} aims to add \textbf{d} to feature maps \textbf{F}, which can be described as:

\begin{equation}
	\label{data_merge}
	\textbf{f}'_i=\textbf{f}_i + p \cdot \textbf{d}_j,
\end{equation}
where, $\mathbf{d}_j$ is the $j$th ($j=1, 2,  \cdots, B$) channel of the input secret data \textbf{d}, $\mathbf{f}_i$ is the $i$th ($i=1, 2,  \cdots, N$) channel of  feature maps, and $p$ is a parameter automatically learned to adjust the strength of merging (see Fig.\ref{fig2} $p$).  "$+$" denotes the mathematical addition operation. $\mathbf{d}_j$ and $\mathbf{f}_i$ are of the same height and width with output stego images. The subscript $i=j+kB$, $i\in [1,N]$, $j\in [1,B]$ and $k$ is a non-negative integer.

When random noise \textbf{n} is input, the data merging operation shrinks to pixel-wise addition between \textbf{n} and each feature map $\mathbf{f}_i$. As a result, a cover image without secret data is generated.

Secret data \textbf{d} and noise \textbf{n} may introduce mosaics or defects into the generated images. To mitigate their impacts,  we adapt a low pass filter in the secret block:
\begin{equation}
	\label{filter}
	\mathcal{F}ilter= \frac{1}{64}\cdot [1,3,3,1]^{T} \times [1,3,3,1],
\end{equation}
here, $\times$ denotes the mathematical matrix multiplication.

\subsection{Structure of Extractor}
\label{extractor}
\begin{figure}[t]
	\centering
	\includegraphics[width=\linewidth]{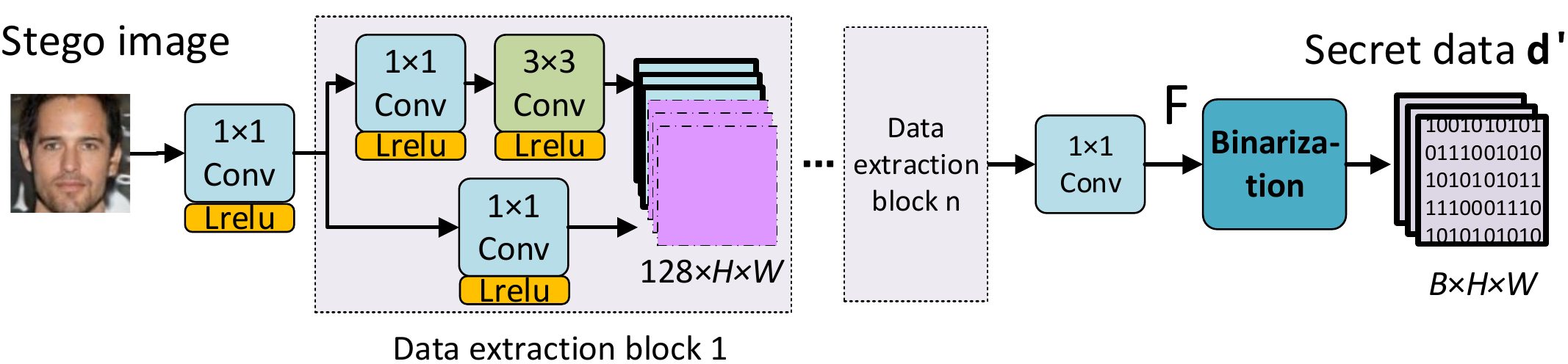}
	\caption{The architecture of proposed data extractor.}
	\label{fig_extr}
\end{figure}

We newly designed a data extractor $E$ to extract the hidden secret data from generated stego images. As shown in Fig.\ref{fig_extr}, it contains several data extraction blocks, two 1×1 convolution layers, and a binarization operation. Each data extraction block includes  some convolution/Lrelu operations. After \textit{n} ( is set to 3 in our scheme) data extraction blocks, the output result is convolved with $B$ 1×1 kernels to produce feature map $\mathbf{F}$, which owns the same size with input secret data (\textit{B} × \textit{H} × \textit{W}). A binarization operation is applied to $\mathbf{F}$, and $\mathbf{d}'$ is the predicted secret data:
\begin{equation}
	\mathbf{d}'=Round(Sigmoid(\mathbf{F})),
\end{equation}
where $Round(\cdot)$ is the rounding operation.

\subsection{Steganalyzer and Discriminator}
To ensure the steganalysis imperceptibility of the generated stego images, we incorporate \textit{D} and \textit{S} as two discriminators for adversarial training. Specifically,  we adopt the original discriminator of stylegan2 \cite{karras2020analyzing}  to ensure the image's visual quality. To generate statistically indistinguishable stego images in steganalysis detection, we use the steganalysis algorithm SR-net\cite{boroumand2018deep} as our steganalyzer \textit{S}. As shown in Fig.\ref{fig1}, the inputs of $S$ are the synthetic cover and stego images, and the inputs of $D$ are real images and synthetic stego images (or cover images, they have similar results).

\subsection{Hierarchical Gradient Decay}
\label{HGD}
Updating the gradients of generator \textit{G} automatically without constraints will result in significant differences between generated cover/stego images (as shown in Fig.\ref{fig-HGD-compare}), which leads to the failure in steganalysis detection. Therefore, we propose the hierarchical gradient decay (HGD) skill to reduce the differences between generated cover/stego images. As illustrated in line 6 of Algorithm\ref{alg1}, when optimizing $G$ and $E$ to minimize $Loss_E$, the HGD skill is applied in \textit{G}, which reduces the gradients of the generator hierarchically as the resolution of the feature map decreases. As a result, the differences between generated cover/stego images are significantly reduced, and the ability to resist steganalysis detection is improved.  

It works because the HGD skill force \textit{G} to hide secret data in high-frequency image details. Steganalysis tools often use the differences of cover/stego images for classification. Once their differences are minor and untraceable enough, steganalysis algorithms are hard to detect the stego images. We find that the proposed secret block is responsible for generating image details while the earlier layers tend to generate low-frequency signals, like shapes and colors of images. Thus, reducing the earlier layers' gradients of $Loss_E$ forces $G$ to hide the secret in high-frequency image details. As a result, the differences between generated cover/stego images are diminished and more randomly distributed. Thus, our generated cover/stego images have a solid ability for anti-steganalysis detection.

The HGD skill gradually decreases the backward gradients from the final secret block to the first general block. Unlike the learning rate decay schedule that adjusts the gradients according to training steps, the HGD skill regulates the updated gradients of different convolution layers according to the size of the feature map. In particular, for feature maps whose size is $(h, w)$, we replace their gradients by:

\begin{equation}
	\label{eq-HGD}
	\mathcal{\hat{G}}=\dfrac{\mathcal{G}}{\delta^{(log_2(\sqrt{HW})-log_2(\sqrt{hw}))}},
\end{equation}
where $\mathcal{G}$ refers to the original backward gradient, and $ \mathcal{\hat{G}}$ is the updated gradient. $H/W$ is the height/width of output images, and $\delta$ is a hyper-parameter.  $ \mathcal{\hat{G}}$ downgrades as the feature size decreases. 

\section{Experimental Results}
Our GSN is trained on TensorFlow 1.14 with four Nvidia 1080Ti GPU. It is evaluated on datasets CelebA\cite{liu2015deep} and Lsun-bedroom\cite{yu2015lsun}.  Adam is used as the optimizer. We set $\lambda_{1}=\lambda_{2}=1$ for $Loss_{G}$ in Eq.\ref{eq-total-loss-g}; $\alpha=1$, $\beta=10$ for $ Loss_{D}$ in Eq.\ref{eq-loss-D};  $\delta=10$ for the HGD skill in Eq.\ref{eq-HGD}.  Input noise's distribution is $\mathbf{n}\sim N(0, \sigma)$, where $\sigma$ is set to 1 for training and 0.1 for testing. 

Frechet inception distance (Fid) \cite{Fid}, extraction accuracy (Acc) and detection error (Pe)  are used to evaluate the visual quality of generated stego images, secret extraction accuracy and the security of our work, respectively. Lower Fid  means better image quality. Acc is calculated as: 
$
\mathrm{Acc}=\frac{\mathbf{d}\bigodot \mathbf{d}'}{len(\mathbf{d})},
$
where, $\mathbf{d}$ and $\mathbf{d'}$ are the input and extracted secret data.  $\bigodot$ is the element-wise $ XNOR $ operation. Pe is a common indicator to evaluate the undetectability of stego images, which is defined as:
$
\mathrm{Pe}=\mathrm{min}_{P_{FA}}\frac{1}{2}(P_{FA} + P_{MD}),
$
where, $P_{FA}$ and $P_{MD}$ are the false alarm rate and missed detection rate. Pe ranges in [0, 1], and its optimal value is 0.5. When Pe is equal to 0.5, the steganalysis tool cannot distinguish the source of images. All our generated images are saved in PNG format. If the length of input secret data is less than 1 bpp, we can add zeros after it. 

\subsection{Performance of Proposed GSN}
We evaluate our GSN with metrics Acc, Fid and Pe. Different secret payloads are obtained by changing the size of input secret data \textbf{d} ($B \times H \times W$), i.e., vary the value of \textit{B} (1 to 8) with \textit{H} and \textit{W} fixed, as shown in Fig.\ref{fig2}. Different GSN models have been trained from scratch with the same setting apart from the payloads and datasets. Then the three metrics are tested with each trained GSN model. Their best performances are recorded in table \ref{table-performance}. The Pe values are obtained by SR-net\cite{boroumand2018deep} trained individually in each configuration.

\begin{table}[b]
	\centering
	\small
	\renewcommand{\arraystretch}{1.0}
	\setlength\tabcolsep{7pt}
	\caption{The performance of proposed GSN.}
	\label{table-performance}
	\begin{tabular}{lcccccc} 
		\toprule
		Datasets  & bpp (\textit{B})     & 1     & 2     & 4     & 6     & 8      \\ 
		\midrule
		\multirow{3}{*}{\begin{tabular}[c]{@{}c@{}}CelebA \\ 128×128\end{tabular}} & Acc (\%) $\uparrow$  & 97.53 & 81.61 & 70.14 & 61.15 & 59.28  \\
		& Fid $\downarrow$     & 13.29 & 15.17 & 16.21 & 16.83 & 18.16  \\
		& Pe $\rightarrow$ 0.5 & 0.479 & 0.499 & 0.501 & 0.502 & 0.498  \\ 
		\midrule
		\multirow{3}{*}{\begin{tabular}[c]{@{}c@{}}Bedroom\\256×256\end{tabular}}  & Acc (\%) $\uparrow$  & 97.25 & 83.19 & 72.13 & 64.17 & 60.94  \\
		& Fid $\downarrow$     & 13.21 & 14.56 & 15.77 & 16.89 & 18.80  \\
		& Pe $\rightarrow$ 0.5     & 0.500 & 0.499 & 0.502 & 0.488 & 0.499  \\
		\bottomrule
	\end{tabular}
\end{table}

\begin{figure}[htb]
	\centering
	\includegraphics[width=1\linewidth]{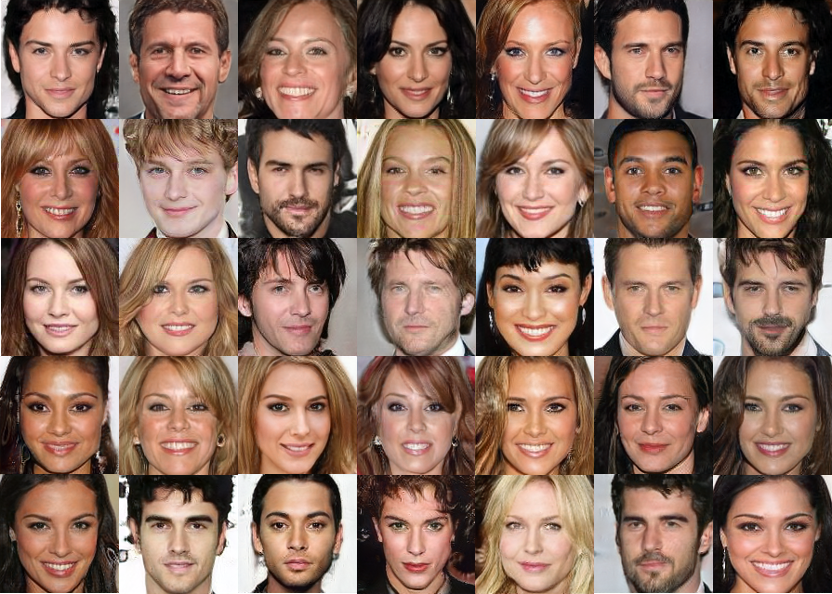}
	\caption{128×128 stego images of faces with various payloads. Images from top to bottom row are with the payload of 1 bpp (bits per pixel), 2bpp, 4bpp, 6bpp and 8bpp.}
	\label{fig-image-faces}
\end{figure}

\begin{figure}[htb]
	\includegraphics[width=1\linewidth]{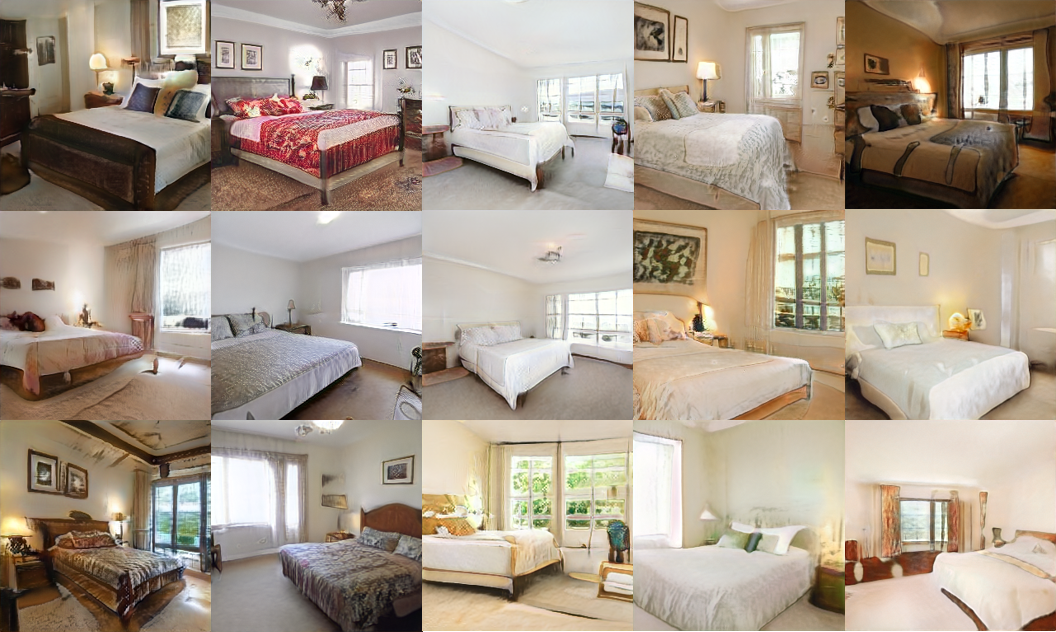}
	\caption{ 256×256  stego images of bedrooms with various payloads. Images from the leftmost to the rightmost column are with the payload of 1 bpp, 2bpp, 4bpp, 6bpp and 8bpp.}
	\label{fig-image-bedroom}
\end{figure}

As shown in table \ref{table-performance}, secret data's extraction accuracy (Acc) decreases gradually as the payload increases. When the payload is 1 bpp,  the Acc values are over 97\% for both datasets. We can employ error correction codes further to increase the Acc values in real applications. Our scheme achieves excellent undetectability regardless of payloads, where the Pe values are close to optimal 0.5.
Fig.\ref{fig-image-faces} and Fig.\ref{fig-image-bedroom} give some  stego image samples. These stego images look real and are hard to be distinguished from real ones.

\begin{figure}[ht]
	\centering
	\begin{subfigure}{0.49\linewidth}
		\centering
		\includegraphics[width=\linewidth]{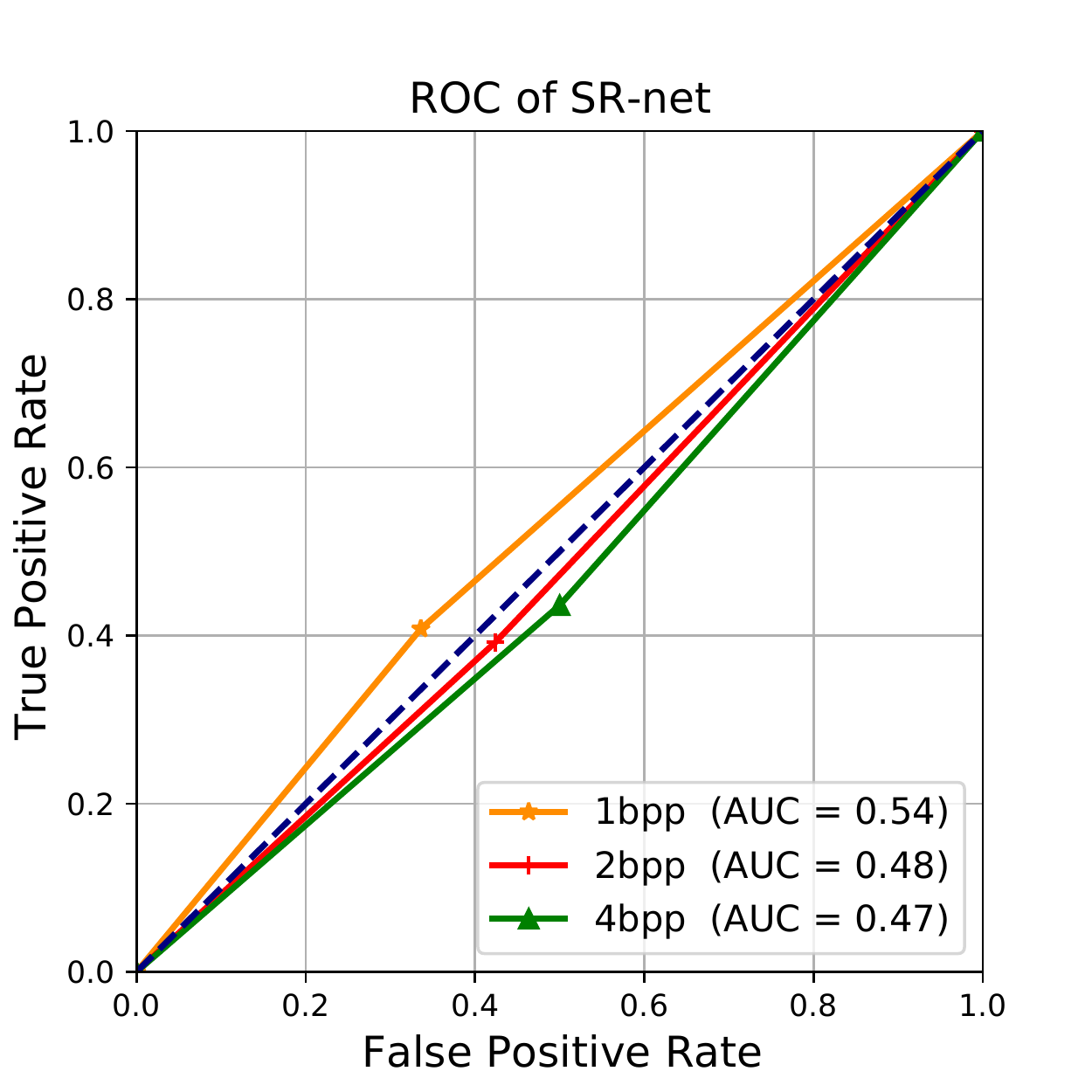}
	\end{subfigure}
	\begin{subfigure}{0.49\linewidth}
		\centering
		\includegraphics[width=\linewidth]{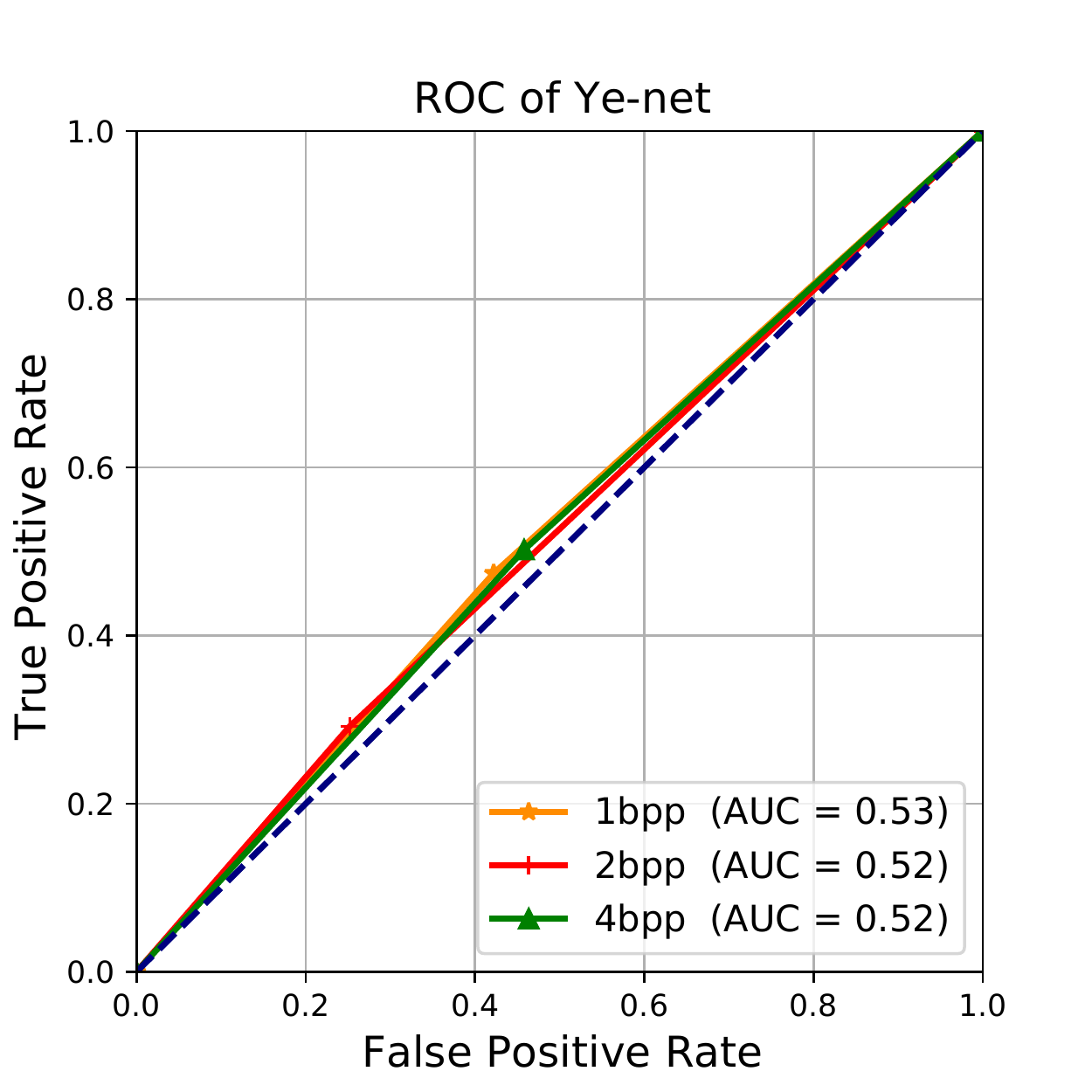}
	\end{subfigure}
	\caption{ROC curves of different steganalysis algorithms.}
	\label{roc}
\end{figure}

\subsection{Security and Steganalysis Resistance}
To verify the security of the proposed method, we test our generated stego images with two advanced steganalysis methods, namely SR-net\cite{boroumand2018deep} and Ye-net\cite{ye2017deep}. Firstly, we trained different GSN models of various payloads on the dataset CelebA. Secondly, we used each GSN model to generate 5000 cover/stego images randomly for steganalysis training and 1000 random cover/stego images for validation and testing. Thirdly, we trained SR-net and Ye-net on each training dataset from scratch, then each well-trained steganalysis model was tested on the corresponding test dataset. At last, the detection ROC curves are plotted in Fig.\ref{roc}. For steganography, the optimal ROC curve is the counter diagonal (dashed line in the figure), and the optimal AUC value is 0.5. As we can see, our ROC curves and AUC values are close to the ideal results, which indicates our proposed GSN is very safe.  

\subsection{Influence of Different Inputs on Stego Image}
Our stego images are generated with two inputs, i.e., latent vector \textbf{z} and secret data \textbf{d} (see Fig.\ref{fig1}). In this section, we analysis their impacts on the generated stego images.

\begin{figure}[htb]	
	\centering
	\begin{subfigure}[]{0.235\linewidth}
		\centering
		\includegraphics[width=1.035\linewidth]{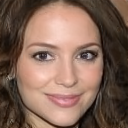}
		\caption{stego 1}
	\end{subfigure}
	\begin{subfigure}[]{0.235\linewidth}
		\centering
		\includegraphics[width=1.035\linewidth]{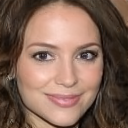}
		\caption{stego  2}
	\end{subfigure}
	\begin{subfigure}[]{0.235\linewidth}
		\centering
		\includegraphics[width=1.035\linewidth]{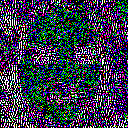}
		\caption{color diff}
	\end{subfigure}
	\begin{subfigure}[]{0.235\linewidth}
		\centering
		\includegraphics[width=1.035\linewidth]{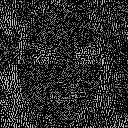}
		\caption{gray diff}
	\end{subfigure}
	\caption{Two stego images generated with different \textbf{d} but the same \textbf{z}. The right two  images show their colored and gray-scaled differences.}
	\label{fig-where-hiding}
\end{figure}

\noindent\textbf{Influence of secret d} ~ We generate two stego images (payload: 1 bpp) with different \textbf{d} but the same \textbf{z}, using a well-trained GSN model. As shown in Fig.\ref{fig-where-hiding}, when \textbf{d} is varied,  these two stego images only differ in image details, such as texture and edges.

\begin{figure}[htb]	
	\centering
	\begin{subfigure}[]{0.235\linewidth}
		\centering
		\includegraphics[width=1.035\linewidth]{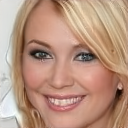}
	\end{subfigure}
	\begin{subfigure}[]{0.235\linewidth}
		\centering
		\includegraphics[width=1.035\linewidth]{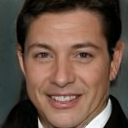}
	\end{subfigure}
	\begin{subfigure}[]{0.235\linewidth}
		\centering
		\includegraphics[width=1.035\linewidth]{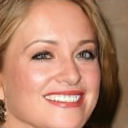}
	\end{subfigure}
	\begin{subfigure}[]{0.235\linewidth}
		\centering
		\includegraphics[width=1.035\linewidth]{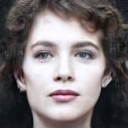}
	\end{subfigure}
	\caption{Four stego images containing the same secret \textbf{d}.}
	\label{fig-same-secret}
\end{figure}

\noindent\textbf{Influence of latent z} ~ Fig.\ref{fig-same-secret} gives four stego images (1 bpp) generated with different \textbf{z} but the same \textbf{d}. By modifying the input latent vector \textbf{z}, the appearances of stego images are changed drastically.

\subsection{Ablation Study}
\begin{table}[b]
	\small
	\renewcommand{\arraystretch}{1.0}
	\setlength\tabcolsep{10pt}
	\centering
	\caption{The performance of different GSN models.}
	\begin{tabular}{@{}lccc@{}}
		\toprule
		\multicolumn{1}{c}{\multirow{2}{*}{Configurations}} & \multicolumn{3}{c}{CelebA,  128×128 }                                        \\
		\cline{2-4}
		\multicolumn{1}{c}{}                                & Fid $\downarrow$  & Acc (\%) $\uparrow$  & Pe $\rightarrow$ 0.5 \\
		\hline
		\ \ Baseline           & 19.45 & 99.23    & 0   \\  
		\ + \ \  Low Pass Filter   & 12.71 & 99.56    & 0  \\
		\ + \ \  Steganalyzer    & 13.03 & 99.67  & 0.035       \\
		\ + \ \ Hierarchical gradient decay  & 13.29  & 97.53    & 0.479 \\
		
		\bottomrule
	\end{tabular}
	\label{table_abliation}
\end{table}

We apply an ablation study to demonstrate the effectiveness of proposed modules and techniques, i.e., the secret block,  steganalyzer, and hierarchical gradient decay (HGD) skill. We rebuild a new GS model without the low pass filter,  steganalyzer and HGD skill, named baseline. To verify the function of the secret block, we cancel the low pass filter (defined in Eq.\ref{filter}) in it. Then these three modules/techniques are added to the baseline one by one cumulatively, and the performance of each new model is demonstrated in Tab.\ref{table_abliation}. All the results are acquired from 128×128 stego images (1 bpp) on CelebA, and the Pe values are detected by SR-net\cite{boroumand2018deep}.

\begin{figure}[]
	\centering
	\includegraphics[width=0.95\linewidth]{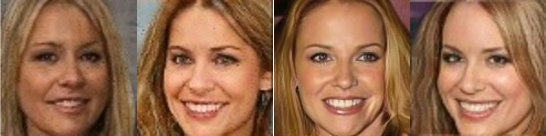}
	\caption{Stego images without (left) / with (right) the low pass filter in secret block.}
	\label{fig5}
\end{figure}

\begin{figure}[]	
	\centering
	\begin{subfigure}[]{0.31\linewidth}
		\centering
		\includegraphics[width=1.03\linewidth]{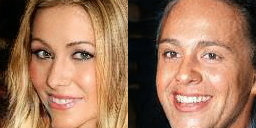}
	\end{subfigure}
	\begin{subfigure}[]{0.31\linewidth}
		\centering
		\includegraphics[width=1.03\linewidth]{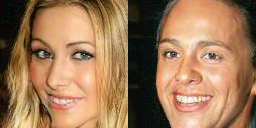}
	\end{subfigure}
	\begin{subfigure}[]{0.31\linewidth}
		\centering
		\includegraphics[width=1.03\linewidth]{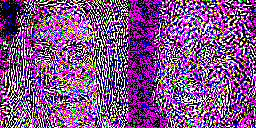}
	\end{subfigure}
	
	\begin{subfigure}[]{0.31\linewidth}
		\centering
		\includegraphics[width=1.03\linewidth]{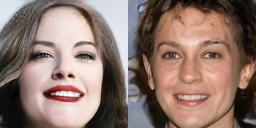}
		\caption{cover images}
	\end{subfigure}
	\begin{subfigure}[]{0.31\linewidth}
		\centering
		\includegraphics[width=1.03\linewidth]{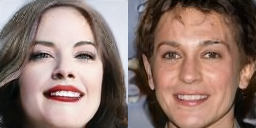}
		\caption{stego images}
	\end{subfigure}
	\begin{subfigure}[]{0.31\linewidth}
		\centering
		\includegraphics[width=1.03\linewidth]{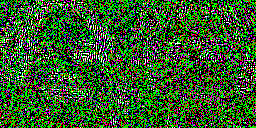}
		\caption{difference × 5}
	\end{subfigure}
	\caption{ Differences between the generated cover/stego images before (top) and after (bottom) using the hierarchical gradient decay skill. The mean absolute error of  cover/stego image pairs is 6.32 (top) and 1.41 (bottom). }
	\label{fig-HGD-compare} 
\end{figure}

By incorporating the low pass filter in the secret block, defects on images are removed, and image quality is enhanced significantly. Fig.\ref{fig5} further demonstrates the image comparison. Adding the steganalyzer will slightly improve the undetectability of stego images with higher Pe. However, it still cannot resist steganalysis detection. The HGD skill improves the Pe value drastically to 0.479 (the ideal value is 0.5), at the cost of reducing Fid and Acc slightly. Fig.\ref{fig-HGD-compare}  compares the differences of cover/stego image pairs before and after using this skill. As we can see, the mean absolute error decreases from 6.32 to 1.41 after employing the HGD skill. Meanwhile, the differences are more randomly distributed in image contents.

\subsection{Comparison with State-of-the-art}
We compare our work with two types of steganographic methods: 1) Traditional steganography (TS) schemes based on deep learning (DL), which need cover images for data embedding. They are termed DL-based TS for short; 2) Generative steganography (GS) works that can generate stego images without cover media, including tailored GS and DL-based GS methods.  

\begin{table}[t]
	\small
	\setlength\tabcolsep{2pt}
	\centering
	\caption{Performance comparison with SOTA  works.}
	\begin{tabular}{llccccc}
		\toprule
		Types &Methods & \begin{tabular}[c]{@{}c@{}}Image\\type \end{tabular}  & \begin{tabular}[c]{@{}c@{}}Secret\\type \end{tabular} &  \begin{tabular}[c]{@{}c@{}}Payload\\(bpp) $\uparrow$ \end{tabular} & \begin{tabular}[c]{@{}c@{}}Acc(\%)\\$\uparrow$ \end{tabular}  & \begin{tabular}[c]{@{}c@{}}Pe\\ $\rightarrow$ 0.5 \end{tabular}  \\
		\midrule
		\multirow{5}{*}{\begin{tabular}[c]{@{}c@{}} DL-based \\ TS  \end{tabular}}
		& Hidden\cite{zhu2018hidden} & natural   & binary   & 1.83e-3 &99.47                                                          & 0.41                                              \\
		& SteganoGAN\cite{zhang2019steganogan} & natural   & binary                  & 1  &99.90                                                          & 0.01                                              \\
		&Hiding-net\cite{img_in_img} & natural & image  &  \underline{24}                                              & 7.16                                                 & 0.03                                                  \\
		&HiNet\cite{hinet} & natural   & image  &  \underline{24}                                          & 56.84                                                & 0.02                                                 \\ 
		&UDH\cite{udh} & natural  & image    &  \underline{24}                                             & 20.34                                                  & 0.01                                                  \\ \midrule
		\multirow{2}{*}{\begin{tabular}[c]{@{}c@{}}Tailored \\ GS \end{tabular}}
		&Wu\cite{wkc2014} & texture   & binary    & 3.28e-2                                               & $\leq$\underline{100}                                                        & ---                                                  \\
		& Li\cite{li2018toward}    & fingerprint   & binary                                                & 1.34e-3 & $\leq$\underline{100}                                                          & ---                                                \\ \midrule
		\multirow{8}{*}{\begin{tabular}[c]{@{}c@{}}\textbf{DL-based }\\ \textbf{GS} \end{tabular}}
		&Liu\cite{liu2017coverless} &natural    & binary                                                & 3.05e-4                                                & 70.56                                                            & 0.48                                           \\
		&Zhang\cite{sss-GAN} & natural      & binary                                              & 3.05e-4                                                & 71.85                                                          & 0.48                                               \\
		&GSS\cite{GSS} & natural   & binary                                                 &  8.8e-2                                                & 63.50                                                          & 0.46                                               \\
		&Hu-1\cite{hu2018novel} & natural    & binary                                                & 1.83e-2                                               & 90.50                                                          & 0.49                                                  \\
		&Hu-2\cite{hu-2} & natural      & binary                                              & 7.32e-2                                               & 91.73                                                          & 0.51                                                  \\
		
		& Our GSN & natural   & binary                                             & 1                                                     & \textbf{97.53 }                                                &  0.51                                              \\
		& Our GSN & natural & binary                                               & \textbf{2}                                                      & 81.61                                                 &  \underline{\textbf{0.50}}                                              \\
		
		\bottomrule
	\end{tabular}
	\label{table-compare2}
\end{table}

\subsubsection{Comparison with DL-based schemes that need cover images}
We compare our GSN with SOTA DL-based TS methods as shown in Tab.\ref{table-compare2}. All these works need cover images for data hiding. For a fair comparison, the 128×128 cover images of CelebA faces generated by our GSN are used as the cover media. Pe values here are obtained by the steganalysis algorithm Ye-net\cite{ye2017deep}. In the table, Hidden\cite{zhu2018hidden} is a novel solution that embeds watermarks in cover images. SteganoGAN \cite{zhang2019steganogan} is a popular scheme that hides binary secret data in cover images. Hiding-net \cite{img_in_img} can hide secret images into a cover image. HiNet\cite{hinet} tries to conceal secret images into the cover images using a revertible network. UDH\cite{udh} is a newly published network for image concealment. In experiments, we only hide one 128×128 secret image in each cover image, carrying a payload of 24 bpp. Their Acc values are the rates of exactly extracted pixels in 3 channels, i.e., $\mathrm{Acc} = num(pixel_{acc}) / (H \times W \times 3) $. Most of the DL-based TS schemes have low Pe values, which indicates modifying cover images to hide secret data  is easily detectable.

\begin{figure}[t]
	\centering
	\begin{subfigure}{0.23\linewidth}
		\centering
		\includegraphics[width=\linewidth]{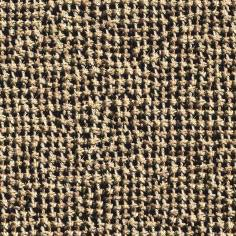}
		\caption{Wu, 32.12}
	\end{subfigure}
	\begin{subfigure}{0.23\linewidth}
		\centering
		\includegraphics[width=\linewidth]{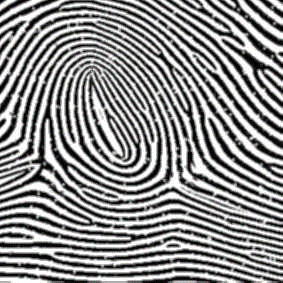}
		\caption{Li, 156.58}
	\end{subfigure}
	\begin{subfigure}{0.23\linewidth}
		\centering
		\includegraphics[width=\linewidth]{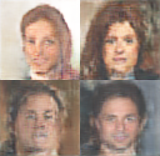}
		\caption{Liu, 56.27}
	\end{subfigure}
	\begin{subfigure}{0.23\linewidth}
		\centering
		\includegraphics[width=\linewidth]{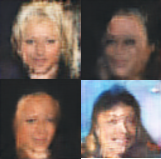}
		\caption{Zhang, 54.54}
	\end{subfigure}
	\begin{subfigure}{0.23\linewidth}
		\centering
		\includegraphics[width=\linewidth]{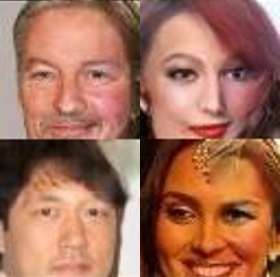}
		\caption{GSS, 45.85}
	\end{subfigure}
	\begin{subfigure}{0.23\linewidth}
		\centering
		\includegraphics[width=\linewidth]{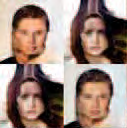}
		\caption{Hu-1, 53.99}
	\end{subfigure}
	\begin{subfigure}{0.23\linewidth}
		\centering
		\includegraphics[width=\linewidth]{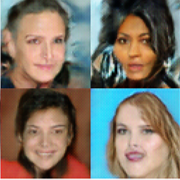}
		\caption{Hu-2, 52.29}
	\end{subfigure}
	\begin{subfigure}{0.23\linewidth}
		\centering
		\includegraphics[width=\linewidth]{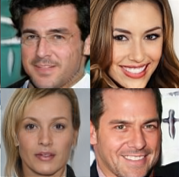}
		\caption{Our, \textbf{13.72}($\downarrow$)}
	\end{subfigure}
	\caption{Stego images generated by different GS methods.  Methods' names and the average Niqe scores are annotated below. Lower Niqe score means better image quality. From (a) to (h), each method carries an absolute payload of 537, 22, 5, 5, 1442, 300, 1200 and 16384 bits/image, respectively.}
	\label{fig-image-compare}
\end{figure}

\subsubsection{Comparison with GS schemes that don't need cover images}
We compare GSN with two SOTA tailored GS schemes as shown in the middle part of Tab.\ref{table-compare2}, where Wu\cite{wkc2014} produces texture stego images and Li\cite{li2018toward} constructs fingerprint stego images. They own low payloads less than 1e-2 but have high Acc values close to 100\%. We also compare our scheme with five SOTA DL-based GS models. They are reimplemented on dataset CelebA, where 128×128 stego images are generated for evaluation. The results are shown in the last rows of Tab.\ref{table-compare2} and Fig.\ref{fig-image-compare}. Both Liu\cite{liu2017coverless} and Zhang\cite{sss-GAN} map secret data to 32 class labels of dataset CelebA. These labels are then input to GANs to generate stego images with the payload of 3.05e-4 bpp (5 bits/image). In  GSS\cite{GSS}, secret data is embedded in corrupted images with a payload of 8.8e-2 bpp, and then stego images are generated by inpainting these images. Both Hu-1\cite{hu2018novel} and Hu-2\cite{hu-2} map secret data to the noise vectors of GANs, with the payloads of 300 bits/image and 1200 bits/image, respectively.

Most DL-based TS methods have poor steganographic security, while our proposed GSN achieves better security performance with higher Pe values. Compared to tailored GS schemes, our GSN can generate realistic natural images with a much higher payload. In DL-based GS schemes, our GSN outperforms the other works overall. The payload is more than 11 times higher than that of theirs. Meanwhile, better Acc and Pe values are obtained. What's more, our work can generate stego images of higher quality. Fig.\ref{fig-image-compare} gives a visual comparison among different GS methods. Results show our stego images are more natural with a lower Niqe score and a higher payload, in which a non-reference image assessor Niqe\cite{niqe} is used to evaluate the image quality.

\section{Conclusions}
This paper proposes a novel GS solution that integrates mutual information mechanism for stego image synthesis. Our  generator is flexibly constructed to generate either cover or stego images, which improves the security in covert communication. We design a delicate secret block to hide secret data into the feature maps during stego image generation, with which high payload and image fidelity are achieved. What's more, a novel hierarchical gradient decay technique is developed to improve the ability of steganalysis resistance. Meanwhile, a discriminator and a steganalyzer are adopted  to improve the visual quality and statistical imperceptibility of generated cover/stego images, respectively. Various experiments have demonstrated the advantages of our GSN over existing works.

\section{Acknowledgment}
This work was supported by the National Natural Science Foundation of China  (U20B2051, 62072114, U20A20178, U1936214).

\newpage
\bibliographystyle{ACM-Reference-Format}
\bibliography{ref}

\end{document}